\documentclass[ai,review,accept,pdftex,moreauthors]{Definitions/mdpi}

\firstpage{1} 
\pubvolume{7}
\issuenum{1}
\articlenumber{6}
\pubyear{2026}
\copyrightyear{2025}
\externaleditor{~}
\datereceived{5 November 2025} 
\daterevised{7 December 2025} 
\dateaccepted{12 December 2025} 
\datepublished{23 December 2025} 
\hreflink{https://doi.org/10.3390/ai7010006} 

\usepackage{longtable}
\usepackage{ragged2e}
\usepackage{makecell}
\newcolumntype{P}[1]{>{\RaggedRight\hspace{0pt}\arraybackslash}p{#1}}
\newcolumntype{C}[1]{>{\centering\arraybackslash}p{#1}}
\usepackage{tikz}
\usetikzlibrary{arrows.meta,positioning,calc,fit,shapes.multipart,backgrounds,matrix}
\pgfdeclarelayer{bg}
\pgfsetlayers{bg,main}

\Title{From Pilots to Practices: A Scoping Review of GenAI-Enabled Personalization in Computer Science Education}

\TitleCitation{From Pilots to Practices: A Scoping Review of GenAI-Enabled Personalization in Computer Science Education}

\Author{Iman 
 Reihanian, Yunfei Hou * and Qingquan Sun}

\AuthorNames{Iman Reihanian, Yunfei Hou and Qingquan Sun}

\AuthorCitation{Reihanian
, I.; Hou, Y.; Sun, Q.}

\address[1]{%
School of Computer Science and Engineering, California State University, San Bernardino, CA 92407
, USA; 008791166@coyote.csusb.edu (I.R.); qsun@csusb.edu 
 (Q.S.)}

\corres{\hspace{-0.82em}Correspondence: yunfei.hou@csusb.edu}

\abstract{Generative AI enables personalized computer science education at scale, yet questions remain about whether such personalization supports or undermines learning. This scoping review synthesizes 32 studies (2023--2025) purposively sampled from 259 records to map personalization mechanisms and effectiveness signals in higher-education CS contexts. We identify five application domains---intelligent tutoring, personalized materials, formative feedback, AI-augmented assessment, and code review---and analyze how design choices shape learning outcomes. Designs incorporating explanation-first guidance, solution withholding, graduated hint ladders, and artifact grounding (student code, tests, and rubrics) consistently show more positive learning processes than unconstrained chat interfaces. Successful implementations share four patterns: context-aware tutoring anchored in student artifacts, multi-level hint structures requiring reflection, composition with traditional CS infrastructure (autograders and rubrics), and human-in-the-loop quality assurance. We propose an exploration-first
 adoption framework emphasizing piloting, instrumentation, learning-preserving defaults, and evidence-based scaling. Four recurrent risks---academic integrity, privacy, bias and equity, and over-reliance---are paired with operational mitigation. Critical evidence gaps include longitudinal effects on skill retention, comparative evaluations of guardrail designs, equity impacts at scale, and standardized replication metrics. The evidence supports generative AI as a mechanism for precision scaffolding when embedded in exploration-first, audit-ready workflows that preserve productive struggle while scaling personalized support.}

\keyword{generative AI; computer science education; personalization; adaptive learning; large language models; feedback; assessment; code review; AI literacy; privacy} 

\begin{document}

\section{Introduction}

Providing timely, personalized support to novice programmers is a long-standing challenge in computer science education. Novice learners often encounter syntax errors or logic bugs that, without~immediate intervention, can lead to frustration and disengagement. While faculties recognize the importance of individualized guidance, delivering it at scale in large introductory courses has historically been impossible due to resource constraints. Generative AI now offers a potential solution to this bottleneck, enabling context-aware assistance that can support students precisely when they are~struggling.

Large language models (LLMs) such as GPT-4/4o and contemporary systems like Claude and DeepSeek have moved from novelty to infrastructure in higher education~\cite{openai2023gpt4,openai2024gpt4o,anthropic2024claude3,deepseek2025r1}. Unlike prior rule-based adaptive systems, LLMs synthesize context-aware explanations, worked examples, and~code-centric feedback on demand, making personalization feasible at the granularity of a single test failure, a~syntax or logic error, or~a misconception articulated in dialogue~\cite{wang2024llmsurvey,xu2024llmsurvey}. At~the same time, GenAI has caused familiar concerns to resurface---integrity, privacy and governance, bias and equity, and~over-reliance---that must be managed
 rather than used to justify paralysis~\cite{used2023ai,educause2024ai,unesco2023guidance,oecd2023deo}.

\begin{enumerate}[label=,leftmargin=0em,labelsep=4mm]
\item[] Policy context.
\end{enumerate}

Institutional responses to GenAI have diverged. Some universities initially prohibited or tightly restricted student use (e.g., Sciences Po's January 2023 ban on ChatGPT---early public release,  
for graded work), while others have adopted enabling, systemwide access models. In~California, the~California State University (CSU) system has deployed \emph{ChatGPT Edu} across all 23 campuses---providing no-cost, single-sign-on access for more than 460{,}000 students and 63{,}000+ employees---and launched the CSU \emph{AI Commons} to centralize training and guidance~\cite{reuters2023sciencespo,csu2025ai,csun2025chatgptedu,csusb2025chatgptedu,campustech2025csu}. This spectrum of policy choices underscores the need for adoption guidance that is practical, evidence-seeking, and~resilient to local~constraints.

\begin{enumerate}[label=,leftmargin=0em,labelsep=4mm]
\item[] {What we mean by personalization.}
\end{enumerate}

We define \emph{personalization} as the \emph{systematic adaptation} of content, feedback, or~task sequencing to a learner's evolving state using observable evidence (code, tests, errors, and dialogue) \cite{xu2024llmsurvey,used2023ai}. In~CS contexts this includes (i) solution-withholding programming assistants that prioritize reasoning; (ii) conversational debugging targeted to concrete error states; (iii) tailored worked examples, Parsons problems, and~practice sets aligned to course context; and (iv) assessment workflows in which LLMs generate tests, rubric-aligned comments, and~explanation-rich grades under human audit~\cite{kazemitabaar2024codeaid,yang2024debuggingaitutor,jury2024workedgen,delcarpio2024personalizedparsons,logacheva2024personalizedexercises,meyer2024llmfeedback,alkafaween2024autograding,xie2024gradelikeahuman}. The~pedagogical aim is \emph{precision scaffolding}: keeping students in productive struggle with stepwise hints, tracing, and~test-driven guidance rather than answer dumps~\cite{kazemitabaar2024codeaid,meyer2024llmfeedback,phung2024gpt4hints}.

\begin{enumerate}[label=,leftmargin=0em,labelsep=4mm]
\item[] A working definition: \emph{exploration-first}.
\end{enumerate}

We use exploration-first
 to denote a deployment stance and workflow in which instructors and institutions \emph{pilot small} \emph{instrument interactions} and~\emph{scale in response to evidence}, with~\emph{learning-preserving defaults} built into tools and policies. Concretely, exploration-first~means:
\begin{enumerate}
  \item \textbf{Help design defaults} that preserve productive struggle: \emph{Explanation-first} hints (pseudocode, tracing, and fault localization), \emph{solution withholding by default}, and~\emph{graduated hint ladders} supported by short \emph{reflection prompts} before escalation.
  \item \textbf{Artifact grounding}: Tutors and feedback are conditioned on the learner's \emph{current code, failing tests, and~assignment specification}; assessment is conducted with \emph{explicit rubrics and exemplars} and \emph{unit tests and mutation checks}.
  \item \textbf{Human-in-the-loop audits} of any generated tests, items, and~grades, with~logs retained for pedagogy and moderation (not ``detector'' policing).
  \item \textbf{Pilot $\rightarrow$ measure $\rightarrow$ scale}: Activation for one section or assignment, examining process and outcome metrics, and~expanding scope when the combined quantitative and qualitative evidence supports doing so. 
  \item \textbf{Enablement governance}: Vetted or enterprise instances, data minimization, and prompt and version change logs; short \emph{allow-lists} in syllabi plus \emph{process evidence} (what was asked, hint levels used, and test history) instead of AI detectors.
\end{enumerate}

\begin{enumerate}[label=,leftmargin=0em,labelsep=4mm]
\item[] Why a scoping review now?
\end{enumerate}

Since 2023, institutions have shifted from ad hoc experimentation to exploration-first adoption: course-integrated assistants that guide rather than answer explicit but enabling policies, faculty development, and~vetted tooling~\cite{educause2024ai,tyton2024timeforclass,crimson2023cs50ai,liu2024cs50ai,stanfordaiteachingguide2024,mittll2024genai,penncetli2024,penn2025aiguidance,duke2025ai}. The~literature is expanding quickly but remains heterogeneous in tasks, measures, and~outcomes; many studies are classroom deployments or mixed-method evaluations. A~\emph{scoping} (rather than effect-size) review is appropriate to map applications and mechanisms and surface signals of effectiveness and risk and~to distill design and governance guidance instructors can use~now.

\begin{enumerate}[label=,leftmargin=0em,labelsep=4mm]
\item[] Objective and gap.
\end{enumerate}

Against this backdrop, there remains a lack of integrative work that focuses specifically on \emph{GenAI-enabled personalization} in higher-education computer science, systematically mapping not just use cases but the underlying mechanisms, reported outcome signals, and~recurrent risks. Existing reviews and position papers typically either address AI in education in general or focus on single tools, courses, or~outcome types, leaving instructors and departments without a consolidated view of how personalization is actually being implemented, under~what conditions it appears to support or hinder learning, and~where the evidence remains thin. The~objective of this scoping review is therefore to synthesize recent empirical work to (i) characterize personalization mechanisms across application areas, (ii) identify the types of process and outcome signals that are reported, (iii) relate these mechanisms and signals to established learning-theoretic constructs, and~(iv) surface gaps and design considerations that can inform both practice and future~research.

\begin{enumerate}[label=,leftmargin=0em,labelsep=4mm]
\item[] Contributions.
\end{enumerate}

Focusing on 2023--2025 in higher-education CS, we~contribute:
\begin{enumerate}
  \item A structured map of how GenAI is used to personalize CS learning, emphasizing mechanisms (explanation-first hints, ladders, and~rubric and test grounding) over brands;
  \item A synthesis of \emph{effectiveness signals} (time-to-help, error remediation, feedback quality, and~grading reliability) and the conditions under which they appear;
  \item A consolidation of risks (integrity, privacy, bias and equity, and~over-reliance) with actionable mitigation;
  \item Design principles and workflow patterns for exploration-first personalization; 
  \item Department and institution guidance for policy, vendor vetting, and~AI-aware assessment~\cite{educause2024ai,used2023ai,fpf2024vetting}.
\end{enumerate}

\begin{enumerate}[label=,leftmargin=0em,labelsep=4mm]
\item[] Research questions.
\end{enumerate}

\begin{enumerate}[label=,leftmargin=3em,labelsep=4mm]
  \item[\textbf{RQ1.}] \textbf{Design and mechanisms:} Which personalization mechanisms (explanation-first help, graduated hints, and code-aware dialogue) are most promising without short-circuiting learning? \cite{kazemitabaar2024codeaid,phung2024gpt4hints,meyer2024llmfeedback,yang2024debuggingaitutor}
  \item[\textbf{RQ2.}] \textbf{Effectiveness conditions:} Under what pedagogical and tooling conditions do GenAI approaches improve learning processes and outcomes (and when do they fail)? \cite{jury2024workedgen,logacheva2024personalizedexercises,gabbay2024moocfeedback}
  \item[\textbf{RQ3.}] \textbf{Risk management:} What recurrent risks accompany personalization, and what mitigation is credible in higher education? \cite{used2023ai,unesco2023guidance,oecd2023deo,mitraise2024securing,fpf2024vetting}
  \item[\textbf{RQ4.}] \textbf{Implementation and assessment:} Which workflows (test and rubric pipelines and process evidence) align personalization with durable learning and fairness? \cite{alkafaween2024autograding,xie2024gradelikeahuman,educause2024ai}
  \item[\textbf{RQ5.}] \textbf{Governance and practice:} How are institutions operationalizing responsible use today (policies, training, and vendor vetting), and what practical guidance follows? \cite{tyton2024timeforclass,educause2024ai,liu2024cs50ai,stanfordaiteachingguide2024,mittll2024genai,penncetli2024,duke2025ai}
  \item[\textbf{RQ6.}] \textbf{Evidence gaps:} What longitudinal and comparative studies are needed (for example, ladder designs and equity impacts)? \cite{oecd2023deo,educause2024ai}
\end{enumerate}

\section{Background and Related~Work}
\unskip

\subsection{From ITSs to~LLMs}
Classic Intelligent Tutoring Systems (ITSs) modeled learner knowledge and errors to deliver stepwise hints and mastery-based sequencing~\cite{carbonell1970,sleemanbrown1982,woolf2009,nwana1990,corbettanderson1994}. Later work refined knowledge tracing and data-driven adaptivity~\cite{piech2015,pelanek2017}. LLMs alter this landscape by \emph{generating} context-specific explanations, examples, and~code-aware feedback through natural-language dialogue~\cite{openai2023gpt4,kasneci2023chatgptgood,xu2024llmsurvey}. The~result is a different granularity of support (line-level commentary and test-oriented guidance) that can be tuned to individual~misconceptions.

\subsection{Clarifying~Terms}
We use \emph{personalization} to denote continuous, evidence-driven tailoring of content, feedback, or~task sequencing. \emph{Adaptation} often refers to real-time adjustments (for example, difficulty and hinting) based on performance signals, whereas \emph{individualization} can include preference or profile-based configuration without continuous updates~\cite{lohr2024adaptive}. Our scope emphasizes code-centric tutoring, targeted feedback, and~sequenced practice that leverage generative models to produce the adapted artifacts themselves~\cite{wang2024llmsurvey}.

\subsection{Affordances Across CS~Subdomains}
Introductory programming affords fine-grained interventions (syntax and logic errors, test failures, tracing), algorithms emphasize strategy explanations and worked examples, and~software engineering invites feedback on design and reviews~\cite{ishaq2024peerj,woolf2009}. Productive scaffolds include explanation-first help, graduated hints, and~practice items aligned to course context and learner~readiness.

\subsection{Pre-GenAI~Baselines}
Before GenAI, personalization drew on behavior and performance signals to deliver immediate feedback, difficulty adjustment, and~sequencing~\cite{rollinson2020aifeedback,lohr2024adaptive}. Autograding pipelines, unit tests, and~program analysis underpinned scalable feedback, but~authored hints and items were costly to produce and maintain. Systematic reviews in programming and medical education foreshadowed GenAI’s promise and limitations~\cite{cavalcanti2021automaticfeedback,lucas2024meded,zawackirichter2019}.

\subsection{GenAI-Enabled Patterns in CS~Education}
Recurring application patterns (2023--2025) include
\begin{enumerate}
  \item \textbf{Solution-withholding assistants and debugging tutors} that deliver explanation-first, context-aware hints~\cite{kazemitabaar2024codeaid,yang2024debuggingaitutor,bassner2024iris,yang2024pensieve,kestin2025aitutor}.
  \item \textbf{Personalized exemplars and practice} (worked examples, Parsons problems, and course-aligned exercises) \cite{jury2024workedgen,logacheva2024personalizedexercises,delcarpio2024personalizedparsons,hou2024codetailor}.
  \item \textls[-15]{\textbf{Targeted formative feedback} with feedback ladders and tutor-style hints~\mbox{\cite{meyer2024llmfeedback,heickal2024feedbackladders,phung2024gpt4hints,gabbay2024moocfeedback,zhu2025feedbot}}.}
  \item \textbf{AI-augmented assessment workflows} (test generation, rubric-guided grading, and~MCQ and exam authoring) \cite{alkafaween2024autograding,xie2024gradelikeahuman,doughty2024mcq,chen2024pythonmcq,hsieh2025aiexams,impey2024autograding,begrading2024,edwards2024autograderllm}.
  \item \textbf{AI-assisted code review} using curated exemplars and model prompts~\cite{lin2024acr,almeida2024aicodereview,shah2021cs1reviewer,cihan2024autocodereview,cihan2025codereviewllm}.
\end{enumerate}

\section{Methods: Scoping~Approach}

We followed Arksey--O'Malley, Levac, and~JBI guidance and~PRISMA-ScR reporting where applicable~\cite{arksey2005scoping,levac2010scoping,peters2020jbi,tricco2018prismascr}. Eligibility was determined according to JBI \emph{population--concept--context (PCC)}: 
\textbf{population} = higher-education CS learners and instructors; \textbf{concept} = GenAI-enabled personalization; \textbf{context} = higher-education CS courses and supports. The~window was 2023--2025. Sources included the ACM Digital Library (SIGCSE TS, ITiCSE, ICER, TOCE), IEEE Xplore (FIE and the ICSE SEIP track), CHI, CSCW, L@S, LAK, EDM, and~indexing via Google Scholar and arXiv. We ran venue-first queries, forward and backward citation chasing, and~hand-searched institutional guidance (policy and governance only).

\begin{enumerate}[label=,leftmargin=0em,labelsep=4mm]
\item[] Registration.
\end{enumerate}

This scoping review was retrospectively registered on the Open Science Framework (OSF); registration details will be provided in the final version
. \url{https://osf.io/bge7y} (accessed on 16 December 2025).

\subsection{Eligibility and~Selection}

Inclusion criteria required that studies (i) took place in higher-education computer science contexts; (ii) implemented or enabled \emph{GenAI-based personalization} rather than generic AI use; (iii) provided empirical evaluation (deployment, design experiment, or~prototype study); (iv) were published in English between 2023 and 2025; and (v) reported sufficient methodological detail to characterize personalization mechanisms. Exclusion criteria eliminated K--12 studies, non-GenAI personalization, opinion pieces, patents, and~papers lacking empirical~evaluation.

Screening proceeded through title/abstract review followed by full-text assessment. Of~the 259 screened records, 59 met full-text eligibility. From~these, we \emph{purposively sampled} 32 studies to support a mechanism-focused scoping synthesis. Purposive sampling is consistent with JBI scoping-review guidance when the goal is to map mechanisms rather than enumerate all instances. Our sampling emphasized analytic suitability rather than outcome~direction.

\begin{enumerate}[label=,leftmargin=0em,labelsep=4mm]
\item[] Rationale for purposive sampling.
\end{enumerate}

A subset of full-text-eligible papers could not meaningfully contribute to mechanism mapping because they lacked operational detail, did not actually implement personalization, or~reported outcomes that were uninterpretable for our analytic aims. We therefore prioritized studies that met all three of the following suitability conditions:

\begin{enumerate}[label=(\alph*)]
  \item \textbf{Mechanism transparency:} Studies that clearly described personalization mechanisms (e.g., hint ladders, explanation-first scaffolding, course-aligned generation, or test- or rubric-grounding) were included. 
  Papers that invoked ``ChatGPT support'' without detailing intervention logic were~excluded.
  
  \item \textbf{Interpretable process or learning outcomes:} Studies reporting measurable learning, debugging, process, or~behavioral outcomes were included. Papers reporting only post hoc satisfaction surveys or generic perceptions without task-linked metrics were excluded because they could not inform mechanism--outcome~relationships.
  
  \item \textbf{Sufficient intervention detail:} Studies that described prompts, constraints, workflows, model grounding, or~tutor policies were included. Excluded papers typically lacked enough detail to map how personalization was implemented (e.g., no description of scaffolding, no explanation of input grounding, or~insufficient reporting of tasks).
\end{enumerate}

\begin{enumerate}[label=,leftmargin=0em,labelsep=4mm]
\item[] Why 27 full-text studies were excluded.
\end{enumerate}

The 27 excluded papers typically exhibited one or more of the following characteristics:

\begin{itemize}
  \item \textbf{Personalization not actually implemented:} The system provided static advice or open-ended chat interaction with no evidence of adaptation.
  \item \textbf{Insufficient mechanism description:} The intervention lacked detail on how hints were generated, how tasks were adapted, or~how the model was conditioned.
  \item \textbf{Outcomes limited to satisfaction surveys:} No behavioral, process, or~learning-related data were reported, preventing mechanism mapping.
  \item \textbf{Redundant or superseded work:} Conference abstracts or short papers from the same research groups that were expanded into more detailed publications were included.
  \item \textbf{Negative or null results with no mechanistic insight:} Some studies reported poor or null outcomes but provided too little detail to attribute failure to design, prompting, scaffolding, or~grounding decisions.
\end{itemize}

\begin{enumerate}[label=,leftmargin=0em,labelsep=4mm]
\item[] Implications for bias.
\end{enumerate}

Because we prioritized mechanism-rich studies, our final corpus likely overrepresents better-specified and more mature deployments. This introduces a known \emph{selection bias} toward successful or interpretable implementations. To~mitigate misinterpretation, we treat outcome patterns as \emph{indicative signals} rather than effectiveness estimates and emphasize throughout that the true distribution of results in the broader literature is likely more~mixed.

\begin{figure}[H]
\resizebox{0.8\textwidth}{!}{%
\begin{tikzpicture}[>=Latex, box/.style={draw, rounded corners, align=center, inner sep=6pt, minimum width=60mm}, arr/.style={-Latex, thick}, node distance=7mm]
\node[box] (id1) {Records identified: venue browsing (SIGCSE, ICER, CHI, CSCW, L@S, LAK, EDM, and~ICSE~(SEIP))\\ \textit{126}};
\node[box, below=of id1] (id2) {Records identified: indexing (Google Scholar and arXiv)\\ \textit{206}};
\node[box, below=of id2] (dup) {Duplicates removed\\ \textit{73}};
\node[box, below=of dup] (scr) {Titles/abstracts screened\\Inclusion: higher education CS and GenAI-personalization\\Exclusion: K--12 only, non-GenAI, opinion\\ \textit{259}};
\node[box, below=of scr] (full) {Full-text assessed for eligibility\\ \textit{59}};
\node[box, below=of full] (inc) {Studies included in synthesis (2023--2025)\\ \textit{32}};
\draw[arr] (id1) -- (id2);
\draw[arr] (id2) -- (dup);
\draw[arr] (dup) -- (scr);
\draw[arr] (scr) -- (full);
\draw[arr] (full) -- (inc);
\end{tikzpicture}%
}
\caption{PRISMA
-style flow for the scoping review. Counts reflect purposive sampling for mechanism-rich, representative~coverage.}
\label{fig:prisma-scope}
\end{figure}
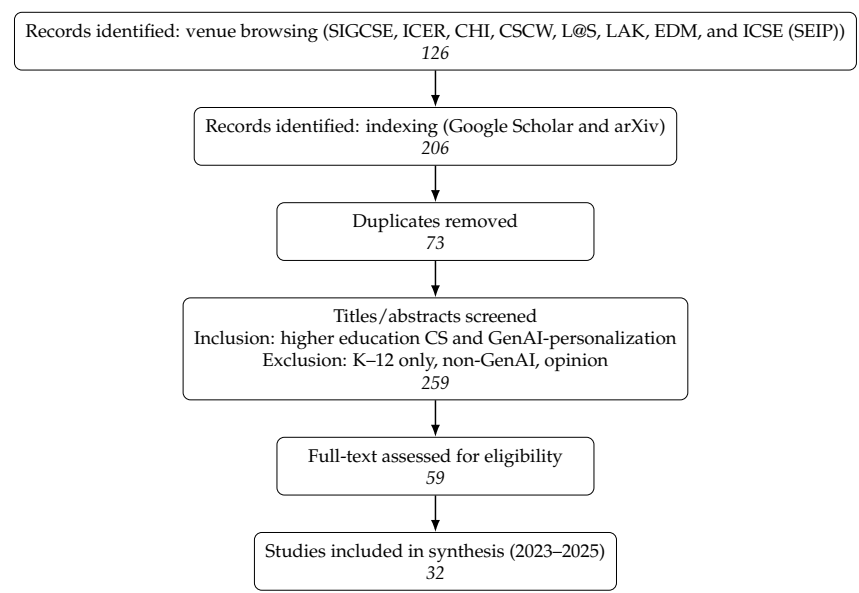

Figure~\ref{fig:prisma-scope} summarizes the screening and purposive sampling process for the included studies.

\subsection{Charting and~Synthesis}
We extracted bibliometrics, instructional context, model/tooling details, personalization mechanisms, implementation, evaluation design, outcomes/effectiveness signals, and~risks/governance. We used a hybrid deductive--inductive synthesis, mapping to an a priori schema (tutoring, learning materials, feedback, assessment, code review, and~governance) and emergent open-coding mechanisms and success/failure conditions. We did not estimate pooled effects, and---consistent with scoping aims and PRISMA-ScR guidance---we did not undertake formal methodological quality appraisal (for example, JBI tools) as~our goal was to map applications and mechanisms rather than to compute aggregate effect-size~estimates.

Table~\ref{tab:taxonomy} presents a taxonomy of GenAI-enabled personalization uses in higher-education computer science, organized by application type, mechanisms, and instructional context.

\begin{table}[H]\renewcommand{\arraystretch}{1.3}
\footnotesize
\caption{Taxonomy
 of GenAI uses for personalization in higher-education CS (illustrative). Abbrev.: LLM = large language model; MCQ = multiple-choice~question.}
\label{tab:taxonomy}
\begin{adjustwidth}{-\extralength}{0cm}
	\begin{tabularx}{\fulllength}{XlXlX}
	
\toprule
\textbf{App. Type} & \textbf{Sources} & \textbf{Personalization Mechanism} & \textbf{Setting and Population} & \textbf{Main Takeaway}\\
\midrule
Solution-withholding programming assistant & \cite{kazemitabaar2024codeaid} & Context-aware explanations, pseudocode, and~line-level annotations that avoid full solutions & Large CS course & Guardrails (no solutions) sustain productive struggle and perceived~learning. \\
\bottomrule
\end{tabularx}
\end{adjustwidth}
\end{table}

\begin{table}[H]\renewcommand{\arraystretch}{1.3}\ContinuedFloat
\footnotesize
\caption{\textit{Cont.}}
\begin{adjustwidth}{-\extralength}{0cm}
	\begin{tabularx}{\fulllength}{XlXlX}
	
\toprule
\textbf{App. Type} & \textbf{Sources} & \textbf{Personalization Mechanism} & \textbf{Setting and Population} & \textbf{Main Takeaway}\\
\midrule
Debugging tutor for novices & \cite{yang2024debuggingaitutor} & Conversational hints grounded in student code and errors & Intro CS & Designs should nudge learners toward strategy-seeking over answer-seeking. \\

Virtual tutor integrated with LMS and IDE & \cite{bassner2024iris} & Tutor role prompts plus context (specifications, code, tests); calibrated assistance & CS course platform & Immediate, personalized support at scale without revealing~solutions. \\
Scalable small-group AI tutoring & \cite{yang2024pensieve} & Group-aware facilitation; targeted prompts and hints & Small-group CS sessions & Personalization extends to group dynamics and~roles. \\
CS61A Bot (course assistant) & \cite{zamfirescu2024cs61abot} & Course-aware assistant supporting task orchestration and help & Large intro CS & Shows feasibility and challenges of course-integrated~assistants. \\

LLM-generated worked examples & \cite{jury2024workedgen} & Course-aligned, level-appropriate exemplars with stepwise explanations & Intro programming & Novices rate examples as useful; curate for~quality. \\
Contextually personalized exercises & \cite{logacheva2024personalizedexercises} & Tailored practice aligned to course context and learner profile & CS courses & Personalized items are viable; quality~varies. \\
Personalized Parsons problems & \cite{delcarpio2024personalizedparsons} & Custom code-rearrangement tasks targeting concepts & CS1 and online practice & Automating generation enables individualized practice at~scale. \\
Personalized Parsons (L@S) & \cite{hou2024codetailor} & Multi-staged, on-demand puzzles that adapt to struggle patterns & CS1 & Engaging support without giving away~solutions. \\

Evidence-based formative feedback & \cite{meyer2024llmfeedback} & Structured, error-specific feedback aligned to pedagogy & Classroom deployments & LLMs surface actionable feedback; prompt design~matters. \\
Feedback ladders for logic errors & \cite{heickal2024feedbackladders} & Graduated hints from high-level cues to specific guidance & Programming assignments & Laddered feedback supports stepwise~progress. \\
Tutor-style hints with validation & \cite{phung2024gpt4hints} & GPT-4 ``tutor'' generates hints; GPT-3.5 ``student'' validates quality & Benchmarks (Python) & Improves hint precision using tests and fixes; test-driven~prompting. \\
Combining LLM + test feedback & \cite{gabbay2024moocfeedback} & LLM feedback complements automated tests in a MOOC & MOOC programming & Hybrid feedback increases correctness and coverage in~practice. \\
FEED-BOT design feedback & \cite{zhu2025feedbot} & Structured, design-recipe-aware formative comments & CS1 design tasks & High-level, structured feedback on design-oriented~tasks. \\

Autograding---test suite generation & \cite{alkafaween2024autograding} & LLM-generated unit tests tuned to specifications & CS1 tasks & Improves coverage; reveals ambiguities; audit~required. \\
``Grade-like-a-human'' pipelines & \cite{xie2024gradelikeahuman} & Rubric-guided, explanation-rich grading with exemplars & Code and short answers & Near-human reliability with explicit rubrics and~calibration. \\
MCQ generation (programming) & \cite{doughty2024mcq,chen2024pythonmcq,hsieh2025aiexams} & Blueprint-aligned MCQs with difficulty control & Intro courses & Acceptable psychometrics with expert~review. \\
AI-authored exams (quality) & \cite{hsieh2025aiexams} & Item generation with validity checks & Web and CS-adjacent & Viable with rigorous vetting~workflows. \\
AI grading and feedback (essays) & \cite{impey2024autograding,burstein2004erater} & Criterion-aligned evaluation with formative comments & General education tasks & Transferable patterns; configure for bias and~accuracy. \\
LLM-supported grading (NCA) & \cite{begrading2024,fcat2013audit} & LLMs for enhanced feedback in programming education & Programming assignments & LLM feedback augments human grading, but there is concern over reliability. \\ 

AI-enhanced code review (ICSE) & \cite{lin2024acr,coverup2024,evogpt2025} & Personalized code critiques learned from review corpora; coverage-guided test pipelines & SE courses & Review quality improves with curated data and~guardrails. \\
AI-assisted code review (JSS) & \cite{almeida2024aicodereview} & Prompted or trained reviewers with exemplars & SE courses & Consistency rises; superficial~suggestions are avoided. \\
Automated code review in practice & \cite{cihan2024autocodereview,cref2024issta} & PR-agent deployed at scale; quality and operations insights & Industrial; SEIP'25 & Industrial lessons for integrating LLM~reviewers. \\
Evaluating LLMs for code review & \cite{cihan2025codereviewllm} & GPT-4o and other models evaluated on correctness and fixes & Benchmarks & LLM reviews help but need human-in-the-loop~processes. \\
Hybrid tutoring/code-review systems & \cite{venugopalan2025lak} & Combining LLMs with existing tutoring intelligence & Informal CS support & Illustrates hybrid human+AI~configurations. \\

	\bottomrule
		\end{tabularx}
	\end{adjustwidth}

\end{table}

\section{Results}

\subsection{Corpus~Characteristics}
Screening and purposive sampling yielded 32
 studies (2023--2025) implementing or enabling personalization in higher-education CS. Most appear in peer-reviewed computing education, HCI, and~software engineering venues (SIGCSE, ITiCSE, ICER, CHI, L@S, LAK, EDM, and~ICSE) with the remainder as detailed preprints. The~modal context is CS1 and CS2 and software engineering courses, with~several large-course deployments (e.g., CS50 and CS61A) \cite{liu2024cs50ai,zamfirescu2024cs61abot}.

\subsection{Application Areas and~Mechanisms}
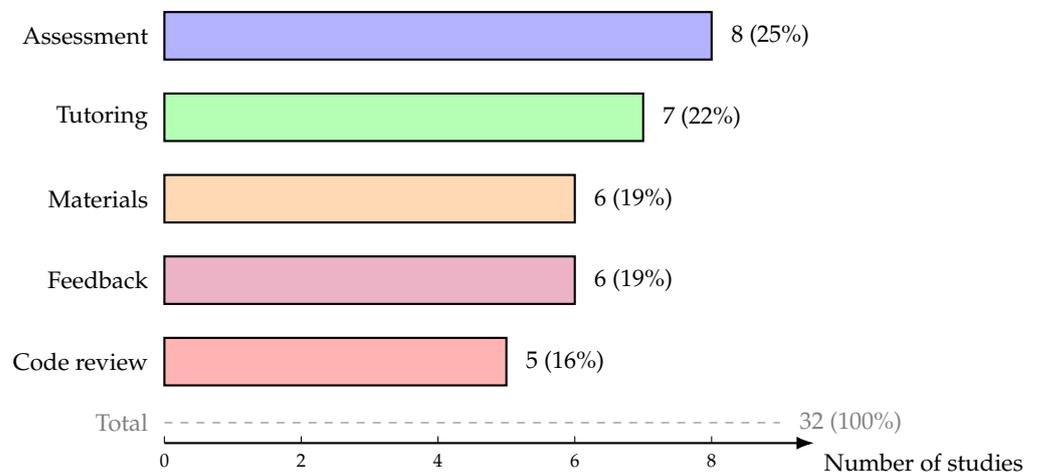
\begin{figure}[H]

\begin{tikzpicture}[x=9mm,y=9mm]
  \draw[thick,-{Latex}] (0,0) -- (9.5,0) node[below right]{\small Number of studies};
  \foreach \x in {0,2,4,6,8} { 
    \draw (\x,0) -- ++(0,0.1) node[below=3pt] {\scriptsize \x}; 
  }
  
  \def\barh{0.7} 
  \def\ysep{1.2}
  
  \node[anchor=east, font=\small] at (-0.1,5*\ysep) {Assessment};
  \draw[fill=blue!30,draw=black,thick] (0,5*\ysep-0.5*\barh) rectangle (8,5*\ysep+0.5*\barh);
  \node[anchor=west, font=\small] at (8.15,5*\ysep) {8 (25\%)};
  
  \node[anchor=east, font=\small] at (-0.1,4*\ysep) {Tutoring};
  \draw[fill=green!30,draw=black,thick] (0,4*\ysep-0.5*\barh) rectangle (7,4*\ysep+0.5*\barh);
  \node[anchor=west, font=\small] at (7.15,4*\ysep) {7 (22\%)};
  
  \node[anchor=east, font=\small] at (-0.1,3*\ysep) {Materials};
  \draw[fill=orange!30,draw=black,thick] (0,3*\ysep-0.5*\barh) rectangle (6,3*\ysep+0.5*\barh);
  \node[anchor=west, font=\small] at (6.15,3*\ysep) {6 (19\%)};
  
  \node[anchor=east, font=\small] at (-0.1,2*\ysep) {Feedback};
  \draw[fill=purple!30,draw=black,thick] (0,2*\ysep-0.5*\barh) rectangle (6,2*\ysep+0.5*\barh);
  \node[anchor=west, font=\small] at (6.15,2*\ysep) {6 (19\%)};
  
  \node[anchor=east, font=\small] at (-0.1,1*\ysep) {Code review};
  \draw[fill=red!30,draw=black,thick] (0,1*\ysep-0.5*\barh) rectangle (5,1*\ysep+0.5*\barh);
  \node[anchor=west, font=\small] at (5.15,1*\ysep) {5 (16\%)};
  
  \draw[dashed, gray] (0,0.3) -- (9,0.3);
  \node[anchor=east, font=\small, gray] at (-0.1,0.3) {Total};
  \node[anchor=west, font=\small, gray] at (9.15,0.3) {32 (100\%)};
\end{tikzpicture}
\caption{Distribution 
 of studies by application area in the included corpus ($N~{=}~32$). Colors distinguish application types for visual~clarity.}
\label{fig:area_bars}
\end{figure}

We classify studies into five non-overlapping application areas (Table~\ref{tab:area_counts}); personalization mechanisms manifest as \emph{explanation-first tutoring and graduated hints}, \emph{course-aligned generation of examples and exercises}, \emph{targeted formative feedback}, \emph{test- and rubric-driven assessment}, and~\emph{AI-assisted code review}. Figure~\ref{fig:area_bars} visualizes the distribution of studies across these application areas.

\begin{table}[H]

\caption{Distribution of included studies by primary application area (non-overlapping).}
\label{tab:area_counts}
\begin{tabular}{P{0.5\linewidth} P{0.30\linewidth} C{0.1\linewidth}}
\toprule
\textbf{Application Area} & \textbf{Representative Sources} & \textbf{\emph{n} (\%)}\\
\midrule
AI-augmented assessment (tests, grading, item and exam generation) & \cite{alkafaween2024autograding,xie2024gradelikeahuman,doughty2024mcq,chen2024pythonmcq,hsieh2025aiexams,impey2024autograding,begrading2024,edwards2024autograderllm} & 8 (25.0)\\
Tutoring and assistants & \cite{kazemitabaar2024codeaid,yang2024debuggingaitutor,bassner2024iris,yang2024pensieve,kestin2025aitutor,zamfirescu2024cs61abot,venugopalan2025lak} & 7 (21.9)\\
Personalized learning materials & \cite{jury2024workedgen,logacheva2024personalizedexercises,delcarpio2024personalizedparsons,hou2024codetailor} & 6 (18.8)\\
Targeted formative feedback & \cite{meyer2024llmfeedback,heickal2024feedbackladders,phung2024gpt4hints,gabbay2024moocfeedback,zhu2025feedbot} & 6 (18.8)\\
AI-assisted code review (SE) & \cite{lin2024acr,almeida2024aicodereview,cihan2024autocodereview,cihan2025codereviewllm} & 5 (15.6)\\
\midrule
\multicolumn{2}{r}{Total
} & 32 (100)
\\
\bottomrule
\end{tabular}
\end{table}
\unskip

\unskip

\subsection{Measures and~Constructs}
Evaluation constructs included time-to-help; error remediation (fraction of failing tests resolved and next-attempt correctness); perceived understanding and utility; feedback quality (specificity, actionability, and~alignment; inter-rater $\kappa$); grading reliability (QWK or adjacent agreement); test quality (statement or branch coverage, mutation score, and~unique edge cases); item and exam quality (difficulty, discrimination, KR-20, or $\alpha$); help-seeking behavior (hint versus solution requests and escalation); instructor and TA effort (authoring and audit time); and developer metrics in code review (precision and recall of issue detection and review acceptance). See Table~\ref{tab:measures}.

\begin{table}[H]
\renewcommand{\arraystretch}{1.3}
\caption{Common evaluation constructs and typical~operationalizations.}
\label{tab:measures}
	\begin{adjustwidth}{-\extralength}{0cm}
		\begin{tabularx}{\fulllength}{lXl}
\toprule
\textbf{Construct} & \textbf{Operationalization} & \textbf{Sources}\\
\midrule
Time-to-help & Latency from request to first actionable hint (IDE or LMS logs). & \cite{kazemitabaar2024codeaid,yang2024debuggingaitutor,nielsen1993responsetimes,akoglu2014soresponsetime,piazza2011press,ucdavis2021piazzastats,tojned2019boards} \\
Error remediation & Share of failing tests resolved; next-attempt correctness; debug step count. & \cite{yang2024debuggingaitutor,alkafaween2024autograding,coverup2024,evogpt2025,cref2024issta} \\
Perceived understanding and utility & Post-task Likert on clarity, usefulness, and~confidence; coded rationales. & \cite{jury2024workedgen,logacheva2024personalizedexercises,prather2024wideninggap,zvielgirshin2024noviceai,pew2024teachers} \\
Feedback quality & Rubric-coded specificity, actionability, and~alignment; inter-rater agreement ($\kappa$). & \cite{meyer2024llmfeedback,heickal2024feedbackladders,phung2024gpt4hints,price2017hintquality,landiskoch1977,roll2006helptutor} \\
Grading reliability & QWK, Pearson or Spearman $r$, exact or adjacent agreement. & \cite{xie2024gradelikeahuman,begrading2024,burstein2004erater,fcat2013audit} \\
Test and coverage quality & Statement or branch coverage; mutation score; unique edge cases surfaced. & \cite{alkafaween2024autograding,coverup2024,evogpt2025} \\
Item and exam quality & Difficulty ($p$), discrimination (point-biserial), KR-20 or $\alpha$; expert review. & \cite{doughty2024mcq,chen2024pythonmcq,hsieh2025aiexams} \\
Help-seeking behavior & Proportion of hint versus solution requests; escalation; prompt taxonomy counts. & \cite{yang2024debuggingaitutor,roll2006helptutor,price2017hintquality,rahe2025howstudentsusegenai} \\
Instructor and TA effort & Authoring, curation, and~audit time; TA workload deltas; review pass rates. & \cite{doughty2024mcq,liu2024cs50ai,harvardcs50ai2023,crimson2023cs50ai} \\
Code-review efficacy & Precision and recall of true issues; fix acceptance; developer effort. & \cite{lin2024acr,cihan2024autocodereview,cihan2025codereviewllm} \\
\bottomrule
\end{tabularx}
	\end{adjustwidth}
\end{table}

\subsection{Descriptive Outcome~Signals}
\textbf{Tutoring/assistants.} Classroom deployments and observations report faster help and higher perceived understanding when assistance emphasizes explanation, pseudocode, and~staged hints while withholding final solutions by default; group-aware facilitation is feasible; and unguarded chat drifts toward answer-seeking~\cite{kazemitabaar2024codeaid,yang2024debuggingaitutor,bassner2024iris,yang2024pensieve,kestin2025aitutor,rahe2025howstudentsusegenai}.

\textbf{Personalized materials.} LLM-generated worked examples and practice sets are often rated usable and helpful by novices; quality varies and benefits from instructor review; and on-demand Parsons puzzles can adapt to struggle patterns~\cite{jury2024workedgen,logacheva2024personalizedexercises,delcarpio2024personalizedparsons,hou2024codetailor}.

\textbf{Targeted feedback.} Structured, error-specific feedback and feedback ladders improve perceived clarity and actionability; tutor-style hints benefit from test/fix grounding and quality validation; hybrid LLM plus test feedback in MOOCs is promising; and design-oriented formative feedback is emerging~\cite{meyer2024llmfeedback,heickal2024feedbackladders,phung2024gpt4hints,gabbay2024moocfeedback,zhu2025feedbot}.

\textbf{Assessment.} LLM-generated unit tests can increase coverage and surface edge cases and ambiguities; rubric-guided grading pipelines can approach human-level agreement when explicit rubrics and exemplars, plus calibration, are used; and MCQ and exam generation is viable with expert review and vetting workflows~\cite{alkafaween2024autograding,xie2024gradelikeahuman,doughty2024mcq,chen2024pythonmcq,hsieh2025aiexams,impey2024autograding,begrading2024,edwards2024autograderllm}.

\textbf{Code review.} Models trained or prompted with high-quality review corpora produce more consistent, personalized critiques; industrial deployments highlight value and pitfalls; and human-in-the-loop processes 
remain essential~\cite{lin2024acr,almeida2024aicodereview,cihan2024autocodereview,cihan2025codereviewllm}.

\section{Comparative Analysis: Design Patterns and~Outcomes}
\label{sec:comparative}
\unskip

\subsection{Design Pattern~Effectiveness}
We coded each study by its primary design pattern and the overall outcome valence as reported by the authors (for example, positive, mixed, or~negative with respect to stated goals). Recurring patterns included explanation-first, solution-withholding tutoring; graduated hint ladders; test- and rubric-grounded assessment; course-aligned generation of examples and exercises; and unconstrained chat~interfaces.

Across the corpus, explanation-first and solution-withholding designs, graduated hints, and~test- and rubric-grounded assessment were \emph{consistently associated} with more positive reported outcomes than unconstrained chat interfaces. Unconstrained chat, especially when it routinely produced complete solutions, appeared more frequently in studies describing mixed or weaker learning benefits, concerns about over-reliance, or~integrity~risks.

Table~\ref{tab:patterns} summarizes design patterns and high-level observations without attempting formal quantitative~comparison.

\begin{table}[H]
\renewcommand{\arraystretch}{1.3}
\caption{Design patterns observed across the 32 studies and common outcome~themes.}
\label{tab:patterns}
	\begin{adjustwidth}{-\extralength}{0cm}
		\begin{tabularx}{\fulllength}{llX}
\toprule
\textbf{Design Pattern} & \textbf{Typical Outcome Tone} & \textbf{Common Observations} \\
\midrule
Explanation-first + solution withholding & Often positive & Supports productive struggle and perceived learning; requires AI literacy to discourage answer-seeking. \\
Graduated hint ladders & Often positive & Aligns with stepwise scaffolding; development cost and tuning are non-trivial. \\
Test/rubric-grounded assessment & Often positive & Reliability improves when coupled with clear rubrics, exemplars, and~audit; hallucinations surface when specs are vague. \\
Course-aligned generation (examples, exercises) & Often positive & Helps with practice at scale; quality variance highlights need for instructor review and curation. \\
Unconstrained chat interface & Often mixed or negative & Solution dumping, reduced productive struggle, integrity concerns, and~over-reliance are recurrent issues. \\
\bottomrule
\end{tabularx}
	\end{adjustwidth}
\end{table}

\subsection{Condition~Analysis}
Studies reporting positive outcomes commonly share several implementation~conditions:
\begin{enumerate}
\item \textbf{Artifact grounding}: Tutoring and feedback anchored in students' current code, failing tests, and~assignment specifications.
\item \textbf{Quality assurance loops}: Human review of generated tests, items, hints, or~grades before or alongside student exposure.
\item \textbf{Graduated scaffolding}: Multi-level hint structures or feedback ladders requiring reflection or effort before escalation.
\item \textbf{AI literacy integration}: Explicit instruction on effective help-seeking, limitations of tools, and~expectations around academic integrity.
\end{enumerate}

Conversely, studies reporting mixed or negative outcomes often~exhibit the following:
\begin{itemize}
\item Unconstrained access to solutions early in the interaction;
\item Grading prompts without explicit rubrics or exemplar calibration;
\item Limited or no instructor review of generated content;
\item Weak integration with existing course infrastructure (autograders, LMS, and~version control).
\end{itemize}

\subsection{Mechanism--Outcome~Mapping}
\begin{figure}[H]

\begin{tikzpicture}[
  node distance=1.3cm and 1.5cm,
  box/.style={rectangle, draw, rounded corners, align=center, minimum width=2.8cm, minimum height=0.9cm, font=\small},
  success/.style={box, fill=green!20},
  risk/.style={box, fill=red!20},
  mech/.style={box, fill=blue!10},
  cond/.style={box, fill=yellow!15},
]

\node[mech] (xpl) {Explanation-first\\guidance};
\node[mech, right=of xpl] (ladder) {Graduated\\hint ladders};
\node[mech, right=of ladder] (ground) {Test and rubric\\grounding};

\node[cond, below=of ladder] (conds) {Implementation:\\Artifact grounding\\Quality loops\\AI literacy};

\node[success, below=of conds] (outcomes) {Positive patterns:\\faster help, clearer\\feedback, improved\\learning processes};

\node[risk, right=3cm of outcomes] (anticond) {Anti-patterns:\\Solution dumping\\No rubrics\\No review};

\node[box, below=of anticond, fill=red!10] (negout) {Negative patterns:\\over-reliance, integrity\\concerns, uneven benefit};

\draw[-Latex, thick] (xpl) -- (conds);
\draw[-Latex, thick] (ladder) -- (conds);
\draw[-Latex, thick] (ground) -- (conds);
\draw[-Latex, thick] (conds) -- (outcomes);
\draw[-Latex, thick, dashed] (anticond) -- (negout);

\end{tikzpicture}
\caption{Conceptual 
 relationships between design mechanisms, implementation conditions, and~outcome patterns. Blue boxes denote design mechanisms; yellow boxes denote implementation conditions; green boxes denote patterns associated with positive learning outcomes; red boxes denote risk and negative outcome patterns. Solid arrows indicate patterns associated with positive outcomes; dashed arrows indicate patterns associated with risks and mixed~outcomes.}
\label{fig:mechanism-outcome} 
\end{figure}
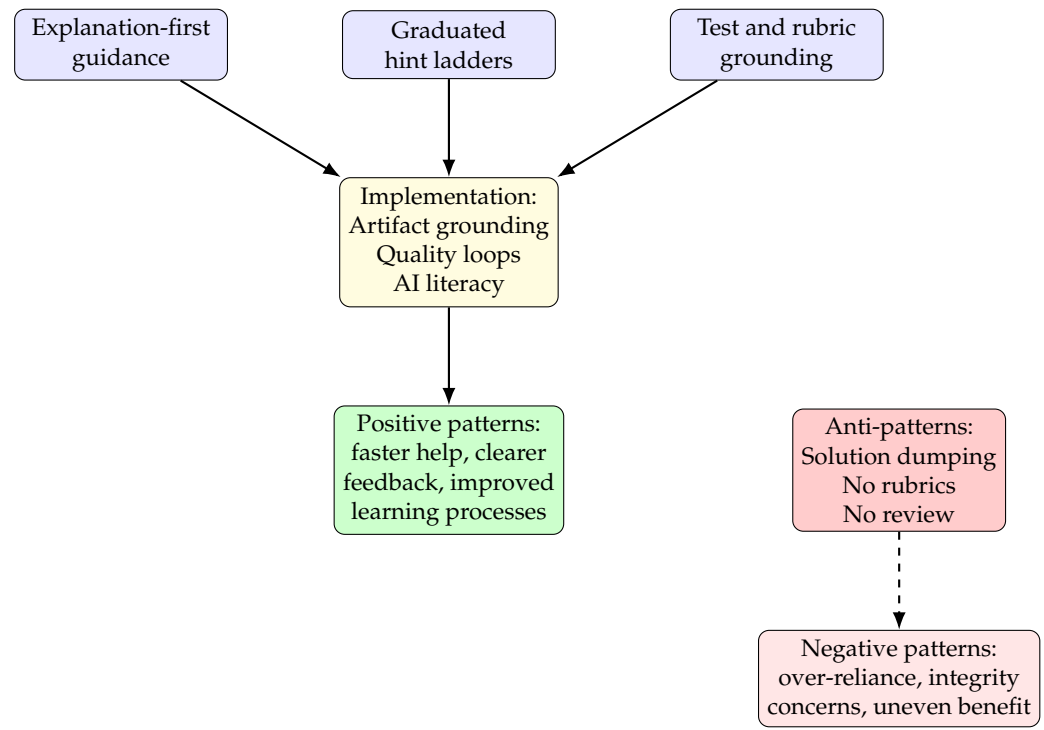

Figure~\ref{fig:mechanism-outcome} visualizes relationships between personalization mechanisms, implementation conditions, and~learning outcomes at a conceptual~level.

\unskip

\section{Discussion}
\unskip

\subsection{RQ1: Design~Mechanisms}
\label{sec:rq1}

Four design patterns recur across the corpus. First, \emph{explanation-first, solution-withholding assistance} prioritizes tracing, error localization, and~pseudocode over finished answers~\cite{kazemitabaar2024codeaid,yang2024debuggingaitutor}. Second, \emph{graduated feedback ladders} escalate from strategy-level cues to line-level guidance~\cite{meyer2024llmfeedback,heickal2024feedbackladders}. Third, \emph{course-aligned generation} of examples and exercises (worked examples and Parsons problems) tailors content to local curricula~\cite{jury2024workedgen,hou2024codetailor}. Fourth, \emph{test- and rubric-driven pipelines} ground assessment and tutor-style hints in explicit specifications and criteria~\cite{alkafaween2024autograding,xie2024gradelikeahuman,phung2024gpt4hints}. Anti-patterns include unconstrained chat that yields full solutions early and~grading prompts without explicit rubrics and~calibration.

These mechanisms are consistent with earlier work on ITSs, help-seeking, and~feedback quality, where granularity of support, alignment with learner state, and~clarity of criteria matter at least as much as raw system accuracy~\cite{carbonell1970,woolf2009,roll2006helptutor,price2017hintquality}.

\subsection{RQ2: Effectiveness~Conditions}
\label{sec:rq2}

Improvements tend to arise under conditions that include guardrails preserving student effort (solution withholding and reflection between hint levels); tight task grounding (student code, tests, and~specifications); structured evaluation artifacts (unit tests, mutation checks, and~rubrics); human-in-the-loop curation (items and examples); and AI literacy plus process evidence to incentivize strategy-seeking~\cite{kazemitabaar2024codeaid,alkafaween2024autograding,xie2024gradelikeahuman,educause2024ai,qaa2023assessmentai}.

Failure modes include answer-seeking drift, variable output quality without review, rubric-poor grading prompts, and~equity risks from uneven access and over-reliance~\mbox{\cite{yang2024debuggingaitutor,logacheva2024personalizedexercises,prather2024wideninggap,zvielgirshin2024noviceai}}. These conditions mirror broader findings on adaptive learning systems and persuasive technology: design choices shape not only learning outcomes but help-seeking habits and motivational dynamics~\cite{lohr2024adaptive,fogg2003persuasive,bandura1977slt,deciryan2000sdt}.

\subsection{RQ3: Risks and~mitigation}
\label{sec:rq3}

\emph{Integrity and over-reliance} can be mitigated via solution withholding, oral defenses or code walkthroughs, process evidence (commit histories and prompt logs), and~a mix of AI-permitted and AI-free assessments~\cite{qaa2023assessmentai,jisc2024assessflow,harvardcs50ai2023}. \emph{Privacy and governance} are addressed via vendor vetting (retention, training use, and~access controls), enterprise instances, data minimization, and~consent pathways~\cite{fpf2024vetting,fpf2024heai,mitraise2024securing,educause2024ai}. \emph{Bias and equity} concerns are mitigated when institutions provision licensed access, review outputs for bias, accommodate multilingual learners, and~avoid unreliable detectors~\cite{oecd2024equityai,openai2023classifier,liang2023detectorbias,baker2022biased,oecd2023biaschapter}. \emph{Quality and hallucination} risks are reduced by composing LLMs with tests and rubrics, calibrating prompts, and~auditing outputs and versions~\cite{alkafaween2024autograding,xie2024gradelikeahuman,edwards2024autograderllm}.

\subsection{RQ4: Workflows That Align with Durable~Learning}
\label{sec:rq4}

Three workflow families stand out: (a) \emph{Tutoring:} defaults to explanation-first; ladder hints; requires reflection between levels; throttles or justifies any code emission, aligning with work on help-seeking, desirable difficulties, and~cognitive load~\cite{roll2006helptutor,price2017hintquality,bjork2011,sweller1988clt}. (b) \emph{Assessment:} specifications feed LLM-generated tests, followed by instructor mutation/coverage audit; rubrics guide exemplar-calibrated grading with moderation~\cite{alkafaween2024autograding,xie2024gradelikeahuman,coverup2024,evogpt2025}. (c) \emph{Process-based assessment:} grading design rationales, test-first thinking, and~revision quality; using Viva or code reviews to assess authorship and understanding~\cite{qaa2023assessmentai,jisc2024assessflow}.

\textls[-15]{Operational transparency---publishing model, prompt, and~policy details; logging for pedagogy and audit; and piloting before scale---supports reliability and trust~\mbox{\cite{educause2024ai,educause2024actionplan,liu2024cs50ai,stanfordaiclassroom2024,mittll2024genai,dukeli2024ai,penncetli2024,penn2025aiguidance,wef2024learning}.}}

\subsection{RQ5: Institutional~Practice}
\label{sec:rq5}

Institutions are converging on \emph{policy-backed, centrally supported} adoption: AI-use statements on syllabi, faculty development, vetted tools, and~privacy-preserving defaults~\cite{educause2024ai,tyton2024timeforclass,educause2024actionplan,unesco2023guidance,wef2024learning}. Large-course exemplars (CS50 and CS61A) illustrate assistants that \emph{guide rather than answer} and embed process-based expectations into course culture~\cite{liu2024cs50ai,harvardcs50ai2023,crimson2023cs50ai,zamfirescu2024cs61abot}. System-level initiatives (e.g., CSU’s ChatGPT Edu deployment and AI Commons; ASU–OpenAI partnership; HKU’s shift from temporary bans to enabling policies) highlight the importance of vendor vetting, training, and~governance~\mbox{\cite{csu2025ai,csun2025chatgptedu,csusb2025chatgptedu,campustech2025csu,sciencespo2023ban,hku2023ban,hku2023shift,asu2024openai}.}

\subsection{RQ6: Evidence~Gaps}
\label{sec:rq6}

Priorities include longitudinal learning effects (retention, transfer, and~potential de-skilling); comparative effectiveness of guardrails (ladder designs, code throttling, and~reflection prompts); equity impacts at scale (stratified analyses by preparation, language, and~access); and shared measures for replication (time-to-help, mutation scores, and~grading agreement thresholds) \cite{zawackirichter2019,alkafaween2024autograding,xie2024gradelikeahuman,oecd2024equityai,prather2024wideninggap}. Many of these needs echo earlier calls from AI-in-education and digital-education policy communities for more educator-centered, equity-aware research~\cite{oecd2023deo,unesco2023guidance,used2023ai,baker2022biased}.

\subsection{Theoretical Grounding of~Findings}
\label{sec:theory}

Having addressed our six research questions empirically, we connect our findings to established learning science principles to understand \emph{why} certain design patterns~succeed.

\begin{enumerate}[label=,leftmargin=0em,labelsep=4mm]
\item[] Desirable difficulties and productive struggle.
\end{enumerate}

The effectiveness of solution-withholding designs (RQ1 and RQ2) connects to \emph{desirable difficulties} theory~\cite{bjork2011}: introducing challenges that require effort during learning can improve long-term retention and transfer, even when they slow initial acquisition. GenAI personalization appears most promising when it maintains challenge within the zone of proximal development~\cite{vygotsky1978} while reducing \emph{unproductive} struggle (environment setup, obscure syntax errors, and tooling friction). Graduated hint ladders operationalize this distinction---they provide just-in-time support for unproductive obstacles while preserving the cognitive engagement needed for schema~construction.

\begin{enumerate}[label=,leftmargin=0em,labelsep=4mm]
\item[] Worked examples and fading.
\end{enumerate}

The success of course-aligned worked examples and Parsons problems reflects worked-example research showing that novices learn effectively from studying solutions before generating them~\cite{sweller1985,renkl2002}. The~key insight is \emph{fading}: progressively reducing support as competence grows, moving from complete examples through partially-completed problems to independent practice. LLMs enable dynamic, individualized fading at finer granularity than cohort-level progressions---adapting to each learner's demonstrated understanding rather than seat~time.

\begin{enumerate}[label=,leftmargin=0em,labelsep=4mm]
\item[] Assessment for learning.
\end{enumerate}

Test-driven tutoring and rubric-guided feedback exemplify \emph{assessment for learning}~\mbox{\cite{black1998,sadler1989}}: formative processes that make success criteria explicit, provide actionable feedback, and~create opportunities for revision. The~effectiveness of test-grounded hints and rubric-anchored grading (RQ4) aligns with the idea that transparency about expectations---paired with specific, timely guidance---supports self-regulation and improvement. GenAI amplifies this by scaling individualized feedback that would be impractical for instructors to provide~manually.

\begin{enumerate}[label=,leftmargin=0em,labelsep=4mm]
\item[] Cognitive load management.
\end{enumerate}

The apparent advantages of artifact-grounded assistance (conditioned on student code, tests, and specifications) over generic tutoring align with cognitive load theory~\cite{sweller1988clt}: learning is optimized when extraneous load is minimized and germane load (effort building schemas) is maximized. Context-aware hints reduce the load of translating generic advice to specific code, freeing working memory for conceptual understanding. Conversely, unconstrained chat that provides complete solutions risks eliminating germane load---the very processing that drives~learning.

Taken together, these connections suggest GenAI personalization is not pedagogically novel so much as a \emph{mechanism} for implementing evidence-based practices at scale. The~central challenge is ensuring designs preserve theoretically grounded features (desirable difficulty, graduated scaffolding, and criterion transparency) rather than optimizing for superficial metrics (task completion speed and satisfaction with solution delivery).

\section{Implementation Roadmap for~Departments}
\label{sec:roadmap}

Based on patterns in successful deployments, we propose a phased approach to GenAI personalization adoption (Table~\ref{tab:roadmap}). A practical instructor deployment checklist, including policy approval and communication requirements, is provided in Appendix~\ref{app:checklist}. The~aim is not to enforce hard thresholds but to encourage routine monitoring of process and outcome metrics and structured decision-making, consistent with digital-education roadmap guidance from organizations such as EDUCAUSE, OECD, UNESCO, and~the World Economic Forum~\cite{educause2024ai,educause2024actionplan,oecd2023deo,unesco2023guidance,wef2024learning}.

\begin{enumerate}[label=,leftmargin=0em,labelsep=4mm]
\item[] Critical decision points.
\end{enumerate}

Key 
 decision points include (1) whether pilot data and stakeholder feedback justify continuing or adjusting the intervention; (2) whether early scaling maintains or erodes benefits; and (3) how policies and tooling should evolve as models and institutional constraints~change.

\begin{table}[H]

\caption{Phased implementation roadmap for~departments.}
\label{tab:roadmap}
	\begin{adjustwidth}{-\extralength}{0cm}
	\begin{tabularx}{\fulllength}{lXX}
\toprule
\textbf{Phase}
 & \textbf{Activities} & \textbf{Illustrative Success Indicators} \\
\midrule
Foundation (Months 1--2) & 
\textbullet~Form
 working group (faculty, IT, legal, students)\newline
\textbullet~Conduct vendor vetting (FPF framework)\newline
\textbullet~Draft policy template with AI-use statements\newline
\textbullet~Identify 1--2 volunteer faculty for pilot &
Policy approved; tools vetted; volunteers trained \\
\midrule
Pilot (Months 3--6) & 
\textbullet~Deploy in 1--2 sections (tutoring OR assessment)\newline
\textbullet~Instrument: log key metrics (e.g., help latency, hint usage, grading agreement)\newline
\textbullet~Weekly check-ins; mid-semester survey\newline
\textbullet~Compare to control (exam scores, retention, equity) &
Data collected; no major incidents; preliminary signals encouraging \\
\midrule
Evaluation (Month 7) & 
\textbullet~Summarize process and outcome metrics (e.g., time-to-help, error remediation, feedback quality, grading reliability)\newline
\textbullet~Analyze equity: stratify by preparation, language, and~other relevant factors\newline
\textbullet~Faculty/student debriefs\newline
\textbullet~Reach a departmental judgment on whether benefits outweigh risks &
Metrics and qualitative feedback suggest pedagogical value without clear harm \\
\midrule
Scale (Months 8--12) & 
\textbullet~Expand to additional courses where pilots show promise\newline
\textbullet~Institutionalize: training, documentation, audits\newline
\textbullet~Continuous monitoring: periodic reviews of usage, outcomes, and~equity &
Sustained performance; quality maintained; instructor capacity built \\
\midrule
Sustain (Year 2+) & 
\textbullet~Annual policy review; vendor re-evaluation\newline
\textbullet~Longitudinal studies (retention, transfer)\newline
\textbullet~Share practices via conferences, consortia &
Durable integration; evidence of learning benefits; community contribution \\
\bottomrule
\end{tabularx}
	\end{adjustwidth}
\end{table}

\vspace{-6pt}

\begin{enumerate}[label=,leftmargin=0em,labelsep=4mm]
\item[] Resource requirements.
\end{enumerate}

Realistic resourcing for a mid-sized CS department (10--15 faculty members, 500--800 students) 
may~include
\begin{itemize}
\item \textbf{Year 1 (pilot)}: 0.25 FTE coordinator; 20--30 h faculty training; tool licensing; 10--15 h/week pilot faculty time.
\item \textbf{Years 2--3 (scale)}: 0.5 FTE coordinator; ongoing training (5--10 h/faculty); audit processes (5--10 h/semester per tool); vendor management.
\item \textbf{Ongoing}: Policy review; assessment validation; longitudinal studies (potentially grant-supported).
\end{itemize}

\section{Limitations}

\begin{enumerate}[label=,leftmargin=0em,labelsep=4mm]
\item[] Temporal and selection bias.
\end{enumerate}

Our 2023--2025 window captures early adoption; designs and models are evolving rapidly. Within~this window, we purposively sampled 32 of 59 full-text-eligible studies to prioritize mechanism transparency and analytic richness. Excluded full-texts commonly relied only on student satisfaction, did not clearly implement personalization, lacked sufficient intervention detail to map mechanisms, duplicated stronger work from the same groups, or~reported negative or null outcomes without actionable mechanistic insight. As~a result, our synthesis emphasizes mechanism-rich, often successful deployments and may underrepresent less well-specified or unsuccessful~attempts.

\begin{enumerate}[label=,leftmargin=0em,labelsep=4mm]
\item[] Publication and outcome bias.
\end{enumerate}

Negative results are underrepresented in the published literature, and~combined with our focus on mechanism-rich studies, this likely leads to an optimistic skew in the available evidence. We therefore present effectiveness signals as indicative rather than definitive and caution that the true distribution of outcomes may include more mixed or negative results than the included corpus~suggests.

\begin{enumerate}[label=,leftmargin=0em,labelsep=4mm]
\item[] Quality appraisal and study design.
\end{enumerate}

Many included sources are conference papers or preprints. Consistent with the aims of a scoping review and PRISMA-ScR guidance, we did not conduct formal methodological quality assessment (for example, JBI tools) and did not attempt to compute pooled effect sizes. Readers should interpret our conclusions as mapping applications, mechanisms, and~reported signals rather than as a formal judgment of study~quality.

\begin{enumerate}[label=,leftmargin=0em,labelsep=4mm]
\item[] Heterogeneity in measurement.
\end{enumerate}

Studies use different metrics, making cross-study comparison difficult. We therefore refrain from cross-study quantitative synthesis and instead rely on narrative descriptions of patterns in reported~measures.

\begin{enumerate}[label=,leftmargin=0em,labelsep=4mm]
\item[] Limited longitudinal data.
\end{enumerate}

Most studies are single-semester deployments. Effects on long-term retention, transfer, and~professional preparation remain~unknown.

\begin{enumerate}[label=,leftmargin=0em,labelsep=4mm]
\item[] Equity analysis gaps.
\end{enumerate}

Few studies stratify by student demographics or prior preparation, limiting equity claims and highlighting the need for equity-focused~research.

\section{Future Work and Research~Priorities}
\label{sec:future}
\unskip

\subsection{Critical Research~Needs} 

\begin{enumerate}[label=,leftmargin=0em,labelsep=4mm]
\item[] Longitudinal studies of learning and skill development.
\end{enumerate}

Cohorts should be tracked over 2--4 years to assess (1) retention of concepts learned with GenAI support versus traditional instruction; (2) transfer to advanced courses and professional practice; (3) potential de-skilling effects (reduced debugging ability and over-reliance on suggestions); and (4) career outcomes (internship acquisition and workplace performance). \emph{Needed design}: Multi-institutional cohort studies with matched controls and standardized assessments at graduation and 1--2 years~post-graduation are required.

\begin{enumerate}[label=,leftmargin=0em,labelsep=4mm]
\item[] Comparative effectiveness trials of guardrail designs.
\end{enumerate}

Randomized controlled trials should be conducted to compare, for~example, (1) hint ladder configurations (two-level vs.\ four-level and reflection prompts vs.\ time delays); (2) code throttling thresholds (no code vs.\ pseudocode vs.\ partial snippets); and (3) artifact grounding strategies (tests-only vs.\ tests + rubrics vs.\ tests + exemplars). \emph{Needed design}: Within-course randomization to ladder variants, standardized outcome measures (error remediation, exam scores, debugging tasks), and replication across~institutions are required.

\begin{enumerate}[label=,leftmargin=0em,labelsep=4mm]
\item[] Equity-focused research.
\end{enumerate}

Stratified analyses and participatory design studies are required. These studies should examine (1) differential effects by prior preparation (AP/IB credit and pre-college coding); (2) language background (multilingual learners and non-native English speakers); (3) disability accommodations (screen reader use, extended time, and captioning); and (4) socioeconomic factors (device access and connectivity). \emph{Needed design}: Purposive oversampling of underrepresented groups, mixed methods combining analytics with interviews, and co-design of tools and policies with students are required~\cite{oecd2024equityai,prather2024wideninggap,baker2022biased}.

\begin{enumerate}[label=,leftmargin=0em,labelsep=4mm]
\item[] Standardized benchmarks and shared datasets.
\end{enumerate}

Community development of (1) benchmark problem sets for tutoring (diverse difficulty, languages, error types); (2) grading rubrics and exemplar sets for assessment studies; and (3) consensus metrics (definitions of time-to-help, error remediation, feedback quality) must be carried out. \emph{Needed infrastructure}: Multi-institutional working groups, open repositories (public GitHub repositories 
 and the Open Science Framework (OSF)), and annual benchmark~efforts are required.

\begin{enumerate}[label=,leftmargin=0em,labelsep=4mm]
\item[] Open-source tool development.
\end{enumerate}

Community-maintained alternatives to commercial tools should be employed: (1) solution-withholding assistants integrated with popular IDEs (VS Code, PyCharm); (2) rubric-guided grading frameworks for common LMS platforms (Canvas and Moodle); and (3) test generation pipelines with mutation-driven quality checks. \emph{Needed investment}: Funding for sustainable development, documentation, and~support is required.

\subsection{Practice~Innovations}

\begin{enumerate}[label=,leftmargin=0em,labelsep=4mm]
\item[] Process-based assessment portfolios.
\end{enumerate}

Courses that assess (1) evolution of prompts (from solution-seeking to strategy-seeking); (2) test-first and revision practices logged in version control; and (3) oral defenses or code walkthroughs demonstrating understanding are key. \emph{Implementation}: Rubrics for process quality, LMS integrations for log capture, and~training for faculty on portfolio~grading should be developed.

\begin{enumerate}[label=,leftmargin=0em,labelsep=4mm]
\item[] Multi-institution collaboratives.
\end{enumerate}

Consortia sharing: (1) vetted prompts and system configurations; (2) audit workflows and quality metrics; (3) assessment items and rubrics; and (4) case studies of policy implementation and incidents should be employed. \emph{Examples}: SIGCSE committee work on GenAI in CS education and regional~collaboratives.

\begin{enumerate}[label=,leftmargin=0em,labelsep=4mm]
\item[] Student co-design and AI literacy curricula.
\end{enumerate}

Participatory design processes engaging students in (1) defining help-seeking norms and tool features; (2) developing peer-to-peer AI literacy workshops; and (3) analyzing their own usage patterns and reflecting on learning strategies should be employed. \emph{Models}: User research methods adapted to CS education contexts and credit-bearing seminars on AI-augmented~learning are required.

\subsection{Policy and Governance~Research}

\begin{enumerate}[label=,leftmargin=0em,labelsep=4mm]
\item[] Vendor vetting and contract negotiation.
\end{enumerate}

Empirical studies of (1) actual data practices versus vendor claims (audits and breach disclosures); (2) effectiveness of DPAs and BAAs in protecting student privacy; and (3) lock-in effects and migration costs should be conducted. \emph{Needed}: Institutional data, legal expertise, and partnership with privacy organizations (FPF, EFF, and CDT) are required \cite{fpf2024vetting,fpf2024heai,mitraise2024securing}.

\begin{enumerate}[label=,leftmargin=0em,labelsep=4mm]
\item[] Labor and instructor impact.
\end{enumerate}

Investigations of (1) changes in instructor workload (time saved on grading vs.\ time spent on tool management and audit); (2) deskilling concerns (TAs losing grading experience and faculty losing assessment design practice); and (3) power dynamics (algorithmic management of teaching and surveillance of instructors) should be carried out. \emph{Methods}: Labor study approaches, critical pedagogy frameworks, and ethnography must be employed.

\begin{enumerate}[label=,leftmargin=0em,labelsep=4mm]
\item[] Long-term institutional case studies.
\end{enumerate}

Multi-year documentation of GenAI personalization adoption at diverse institutions should be developed: (1) policy evolution (from prohibition to enablement to normalization); (2) infrastructure development (procurement, support, and training); and (3) cultural change (faculty attitudes and student expectations). \emph{Design}: Longitudinal ethnography, document analysis, and interviews with administrators and~faculty are required.

\section{Conclusions}

GenAI can deliver \emph{precision scaffolding} in CS education---faster help, clearer targeted feedback, and~scalable assessment support---\emph{when} designs emphasize explanation-first tutoring, graduated hints, and~test- and rubric-driven workflows under human oversight. Unconstrained, solution-forward use risks eroding learning and exacerbating integrity and equity issues. An~exploration-first stance---clear goals, enabling policies, vetted tools, and~routine audits---aligns personalization with durable learning and fairness~\cite{kazemitabaar2024codeaid,alkafaween2024autograding,xie2024gradelikeahuman,educause2024ai}.

\begin{enumerate}[label=,leftmargin=0em,labelsep=4mm]
\item[] Actionable takeaways.
\end{enumerate}

\begin{itemize}
\item \textbf{Design for productive struggle:} Default to solution withholding and laddered hints; require reflection between hint levels~\cite{kazemitabaar2024codeaid,heickal2024feedbackladders,bjork2011,sweller1985}.
\item \textbf{Ground feedback in artifacts:} Anchor guidance in student code, tests, and~specifications; compose LLMs with unit tests and rubrics~\cite{meyer2024llmfeedback,alkafaween2024autograding,phung2024gpt4hints,coverup2024}.
\item \textbf{Assess the process:} Grade design rationales, prompt logs, and~oral defenses; avoid reliance on AI detectors~\cite{qaa2023assessmentai,openai2023classifier,liang2023detectorbias,jisc2024assessflow}.
\item \textbf{Institutionalize enablement:} Provide licensed tools, vendor vetting, privacy-by-design defaults, and~faculty/TA training~\cite{fpf2024vetting,educause2024ai,educause2024actionplan,mittll2024genai,dukeli2024ai,penncetli2024,penn2025aiguidance}.
\item \textbf{Monitor equity:} Provision access for all students; audit outputs for bias; study heterogeneous impacts~\cite{oecd2024equityai,prather2024wideninggap,baker2022biased,wef2024learning}.
\item \textbf{Build the evidence:} Invest in longitudinal and comparative studies with shared metrics~\cite{zawackirichter2019,alkafaween2024autograding,xie2024gradelikeahuman,oecd2023deo}.
\end{itemize}

\vspace{6pt} 


\funding{This 
 material is based upon work supported by the National Science Foundation under Grant No.~2524227.}

\institutionalreview{Not applicable.}

\informedconsent{Not applicable.}

\dataavailability{No new data were created or analyzed in this study. Data sharing is not applicable to this article.} 

\acknowledgments{The authors thank the reviewers for their constructive~feedback.}

\conflictsofinterest{The authors declare no conflicts of~interest.} 


\appendixtitles{yes}
\appendixstart
\appendix

\section{Deployment Checklist for~Instructors}
\label{app:checklist}
\unskip

\subsection{Before~Deployment}
\begin{itemize}[label=$\square$]
\item Policy 
 approved by department and communicated to students;
\item Vendor FERPA/GDPR compliance audit completed;
\item Faculty training conducted (tool features, pedagogical strategies, and risk mitigation);
\item Baseline data collection planned (control sections or pre-deployment metrics);
\item Human-in-the-loop audit workflow defined (who reviews what, when, and~how);
\item Syllabus updated with AI-use statement (allowed tools, permitted uses, and citation requirements);
\item Student AI literacy session scheduled (effective help-seeking, tool limitations, and academic integrity).
\end{itemize}

\subsection{During Pilot (Weekly/Bi-Weekly)}
\begin{itemize}[label=$\square$]
\item Interaction logs reviewed for answer-seeking patterns and over-reliance signals;
\item Student feedback collected (weeks 3, 8, 15: utility, clarity, and concerns);
\item Quality spot-checks (sample generated hints, grades, or~tests; verify accuracy and alignment);
\item Equity monitoring (compare usage and outcomes by subgroup where feasible; investigate disparities);
\item Incident log maintained (errors, hallucinations, inappropriate outputs, and student complaints).
\end{itemize}

\subsection{Evaluation at End of~Pilot}
\begin{itemize}[label=$\square$]
\item Key process metrics (e.g., time-to-help, hint usage, and grading agreement) summarized and interpreted;
\item Evidence of meaningful error remediation and/or improved feedback quality relative to baseline;
\item Grading and test-generation workflows checked for reliability and alignment with rubrics and specifications;
\item No major equity concerns identified in stratified analyses (where data permits);
\item Faculty and student feedback indicates that benefits outweigh burdens or risks.
\end{itemize}

\subsection{Decision~Point}
\begin{itemize}[label=$\square$]
\item \textbf{If evidence is broadly positive,} consider scaling to additional sections or courses with ongoing monitoring;
\item \textbf{If evidence is mixed,} diagnose causes (tool design, instructor preparation, and task alignment), refine, and~re-pilot;
\item \textbf{If major risks are identified,} pause or discontinue use pending remediation; document lessons learned.
\end{itemize}

\subsection{Ongoing (Post-Scale)}
\begin{itemize}[label=$\square$]
\item Periodic review of usage, outcome, and~equity metrics;
\item Annual policy review (update for new use cases, model changes, and regulatory shifts);
\item Vendor re-evaluation (privacy practices, pricing, feature roadmap, and lock-in risks);
\item Longitudinal follow-up where feasible (e.g., retention, transfer, and~downstream course performance);
\item Community contribution (share practices, prompts, and~lessons via conferences or repositories).
\end{itemize}


\PublishersNote{}
\end{document}